\documentclass[11pt,a4paper]{article}
\usepackage[hyperref]{ACL2021}
\usepackage{times}
\usepackage{latexsym}
\usepackage{graphicx}
\usepackage{booktabs}
\usepackage{multirow}
\usepackage{amsmath}

\usepackage{array}

\usepackage{microtype}

\aclfinalcopy % Uncomment this line for the final submission
\def\aclpaperid{3384} %  Enter the aacl-ijcnlp Paper ID here

\newcommand{\enlidist}[3]{$\mathbf{E^{#1}}$ $\mathrm{N}^{#2}$ $\mathrm{C}^{#3}$}
\newcommand{\nnlidist}[3]{$\mathrm{E}^{#1}$ $\mathbf{N^{#2}}$ $\mathrm{C}^{#3}$}

\title{

Embracing Ambiguity: Shifting the Training Target of NLI Models
}

\author{

Johannes Mario Meissner$^\dagger$, Napat Thumwanit$^\dagger$, Saku Sugawara$^\ddagger$, Akiko Aizawa$^{\dagger\ddagger}$ \\
$^\dagger$The University of Tokyo, $^\ddagger$National Institute of Informatics \\
\texttt{\{meissner,thumwanit-n,saku,aizawa\}@nii.ac.jp}
  
}

\date{}

\begin{document}
\maketitle
\begin{abstract}

Natural Language Inference (NLI) datasets contain examples with highly ambiguous labels. While many research works do not pay much attention to this fact, several recent efforts have been made to acknowledge and embrace the existence of ambiguity, such as UNLI and ChaosNLI. In this paper, we explore the option of training directly on the estimated label distribution of the annotators in the NLI task, using a learning loss based on this ambiguity distribution instead of the gold-labels. We prepare AmbiNLI, a trial dataset obtained from readily available sources, and show it is possible to reduce ChaosNLI divergence scores when finetuning on this data, a promising first step towards learning how to capture linguistic ambiguity. Additionally, we show that training on the same amount of data but targeting the ambiguity distribution instead of gold-labels can result in models that achieve higher performance and learn better representations for downstream tasks. 

\end{abstract}

\section{Introduction}

% Introduce topic and previous work

Ambiguity is intrinsic to natural language, and creating datasets free of this property is a hard if not impossible task. Previously, it was common to disregard it as noise or as a sign of poor quality data. More recent research, however, has drawn our attention towards the inevitability of ambiguity, and the necessity to take it into consideration when working on natural language understanding tasks \cite{pavlick-kwiatkowski-2019-inherent, chen-etal-2020-uncertain, nie-etal-2020-learn, swayamdipta-etal-2020-dataset}. This ambiguity stems from the lack of proper context or differences in background knowledge between annotators, and leads to a large number of examples where the correctness of labels can be debated. 

% In this work we focus on the task of natural language inference (NLI), specifically in the form presented by the SNLI and MNLI datasets, where the goal is to classify a pair of sentences (premise and hypothesis) as forming either an entailment, a contradiction, or being neutral to each other.

% Introduce previous works and problem

ChaosNLI \cite{nie-etal-2020-learn} is a dataset created by manually annotating a subset of the SNLI \cite{bowman-etal-2015-large}, MNLI \cite{williams-etal-2018-broad}, and $\alpha$NLI \cite{Bhagavatula2020Abductive} datasets. Each of the total 4,645 samples received 100 annotations. Through this data, they were able to generate a probability distribution over the labels for these samples, which they call the human agreement distribution, with the goal of using it to evaluate the ability of current state-of-the-art models to capture ambiguity. The divergence scores between the model's predicted probability distribution and the true target distribution is computed and compared against random and human baselines. They showed that models trained using gold-labels have very poor performance on the task of capturing the human agreement distribution. 

% Proposed solution

Although this is a promising first step, it remains unclear how to train models with a better understanding of ambiguity, and what tangible benefits we can obtain when actually doing so. 
% We believe that due to the very nature of natural language, ambiguity should be considered not only during evaluation, but also when building training datasets and the models themselves.
In this work, we study the possibility of shifting the training target of models from gold-labels to the ambiguity distribution, a simple and intuitive yet until now unexplored approach in this domain.
We hypothesize that when we finetune a model in this way, we can achieve lower divergence scores in the ChaosNLI benchmark. Further, we believe that it should also bring accuracy improvements in NLI and other downstream tasks. The intuition behind our performance expectations is that an ambiguity distribution offers a more informative and less misleading view on the answer to the task, which allows models to learn more from the same data.
% Furthermore, moving from gold-labels to ambiguity distributions can be regarded as process similar to data cleaning or denoising.

We prepare a trial dataset with ambiguity distributions obtained from available SNLI and MNLI data, and run experiments to confirm our hypotheses. We refer to it as AmbiNLI, but we do not encourage its use in further work. Instead, we encourage the community to follow this direction by performing further data collection in this area.

Our main contributions are showing that 1) models trained on ambiguity can more closely capture the true human distribution, 2) they are able to attain higher accuracy under otherwise equal conditions, and 3) they learn better representations for downstream tasks. We release the code used for these experiments.\footnote{\url{https://github.com/mariomeissner/AmbiNLI}}

% ... while future work may involve crowd-sourcing to obtain much larger amounts of data. As our results suggest, it is possible to benefit from a very rough or approximate ambiguity distribution, as opposed to the precision provided in ChaosNLI. This insight will allow to considerably reduce the cost of obtaining ambiguity data in he future.

\section{AmbiNLI}
\subsection{Available Data}
\begin{table}[]
\small
\centering
% \resizebox{\linewidth}{!}{%
\begin{tabular}{@{}lllrr@{}}
\toprule
\textbf{Dataset}      & \textbf{Split}           & \textbf{Used By} & \textbf{\#Samples} & \textbf{\#Labels} \\ \midrule
\multirow{6}{*}{SNLI} & Train                 & UNLI     & 55,517 & 1r  \\ \cmidrule(l){2-5} 
                      & \multirow{3}{*}{Dev.} & UNLI     & 3,040  & 1r  \\
                      &                       & ChaosNLI & 1,514  & 100 \\
                      &                       & Original & 9,842  & 5   \\ \cmidrule(l){2-5} 
                      & \multirow{2}{*}{Test} & UNLI     & 3,040  & 1r  \\
                      &                       & Original & 9,824  & 5   \\ \midrule
\multirow{3}{*}{MNLI} & \multirow{2}{*}{Dev. M.} & ChaosNLI         & 1,599              & 100               \\
                      &                       & Original & 9,815  & 5   \\ \cmidrule(l){2-5} 
                      & Dev. Mism.            & Original & 9,832  & 5   \\ \bottomrule
\end{tabular}%
% }
\caption{Data with enough information to generate a probability distribution over the labels. The marker ``1r" denotes the fact that there is only one data-point available, but it is a regression label in the [0,1] range, so it can be converted.}
\label{ambiguity-data}
\end{table}

Data containing enough information to reveal ambiguity is relatively scarce. To construct AmbiNLI we generated the label distributions from several sources. Table~\ref{ambiguity-data} details the available data that we have taken into consideration.

\paragraph{SNLI / MNLI.} \label{nli-eval-sets}
Both SNLI and MNLI provide labels assigned by 5 annotators on some subsets of the data (marked ``Original" in Table~\ref{ambiguity-data}). Examples where no human agreement could be reached (no majority) were given a special label (-1) and are commonly filtered out. Although the precision given by 5 labels is much lower than that of the 100 annotations provided in ChaosNLI, we believe that even a rough and inaccurate ambiguity representation is more beneficial than gold-labels only. 

\paragraph{UNLI.}
UNLI \cite{chen-etal-2020-uncertain} presents a subset of SNLI as a regression task, where each example is annotated with a real value in the range [0,1]. Values close to 0 indicate contradiction, and values close to 1 represent entailment.
% We use this data to generate a human agreement distribution as well. 
% It is important to note that due to the nature of the labels in this dataset, the strongest type of ambiguity that we could potentially have in other datasets (disagreement between contradiction and entailment) cannot be properly captured. 
Each entry has one label only, but since it real-valued, it is also possible to extract a distribution from it.
Even though it seems to be a less suitable data source, we do intend to investigate its effectiveness for our purposes.
% Once again, we hypothesized that even if the precision in the ambiguity data generated from UNLI is not nearly as high as in ChaosNLI, it might be beneficial to utilize it nonetheless.

\paragraph{ChaosNLI.}
ChaosNLI provides annotations from 100 humans for 3,113 examples in the development sets of SNLI and MNLI. We will call these subsets ChaosSNLI and ChaosMNLI. In order to allow for comparison with the original paper, we use them for testing only.

\subsection{Creating AmbiNLI}
% First, we make sure to avoid overlap between ChaosNLI and ``Original'' data by removing the samples used by ChaosNLI from the data we will include in AmbiNLI for training.
Original SNLI and MNLI data with 5 annotations can be converted to an ambiguity distribution by simply counting the number of annotations for each label and then scaling it down into probabilities. 
%ChaosNLI used a subset of SNLI dev. and MNLI dev. matched. 
% Since we wish to have no overlap between training and test data, we remove those samples from our dataset.
We make sure to avoid overlap between ChaosNLI and ``Original'' data by removing the samples used in ChaosNLI from the data we will include in AmbiNLI.
In the case of UNLI, we have taken only the 55,517 samples from the training set, so there is no overlap with ChaosNLI. We apply a simple linear approach to convert the UNLI regression value $p$ into a probability distribution $z_{\mathrm{NLI}}$, as
described in the following composed function (its plot can be found in the Appendix \ref{section:conversion-function}):
%depicted in the plot of Figure~\ref{fig:unli-conversion} to obtain the ambiguity distribution.
\begin{equation*}
    z_{\mathrm{NLI}}=
    \begin{cases}
        (0,    2p,   1-2p) & p < 0.5 \\
        (2p-1, 2-2p, 0)    & p \geq 0.5.
    \end{cases}
\end{equation*}

The resulting AmbiNLI dataset has 18,152 SNLI examples, 18,048 MNLI examples, and 55,517 UNLI examples, for a total of 91,717 premise-hypothesis pairs with an ambiguity distribution as the target label.

\begin{table}[]
\centering
\small
\begin{tabular}{@{}lrrrr@{}}
\toprule
\textbf{Data} & \multicolumn{2}{c}{\textbf{ChaosSNLI}} & \multicolumn{2}{c}{\textbf{ChaosMNLI}} \\
\textbf{Metric} & \multicolumn{1}{c}{\textbf{JSD$\downarrow$}} & \multicolumn{1}{c}{\textbf{Acc.$\uparrow$}} & \multicolumn{1}{c}{\textbf{JSD$\downarrow$}} & \multicolumn{1}{c}{\textbf{Acc.$\uparrow$}} \\ \midrule
S/MNLI Baseline       & 0.2379          & 0.7497          & 0.3349          & 0.5566          \\ \midrule
+ AmbiSM Gold            & 0.2307          & 0.7497          & 0.3017          & 0.5660          \\
+ AmbiSM              & \textbf{0.1893} & \textbf{0.7550} & 0.2619          & 0.5810          \\ \midrule
+ AmbiU Gold             & 0.3118          & 0.5878          & 0.3183          & 0.5260          \\
+ AmbiU               & 0.2834          & 0.5964          & 0.2843          & 0.5178          \\
+ AmbiU Filt.            & 0.2302          & 0.6790          & \textbf{0.2231} & 0.5779          \\ \midrule
+ AmbiSMU Gold           & 0.2936          & 0.6162          & 0.3540          & 0.5822          \\
+ AmbiSMU             & 0.2554          & 0.6420          & 0.2575          & 0.5766          \\
+ AmbiSMU Filt.           & 0.2155          & 0.7107          & 0.2748          & \textbf{0.5835} \\ \bottomrule
\end{tabular}
\caption{Main results of our finetuning experiments on AmbiNLI. \emph{Gold} means that gold-labels, and not ambiguity distribution, was used for training. \emph{Filt.} indicates that extreme examples in UNLI have been filtered out.}
\label{divergence-table}
\end{table}

\section{Experiments}

% To study the effect of training on our dataset and validate our hypotheses, we executed a series of experiments. 
In our experiments, we use BERT-base \cite{devlin-etal-2019-bert} with pre-trained weights and a softmax classification head. We use a batch size of 128 and learning rate of 1e-5.

\paragraph{Learning to capture question ambiguity.} \label{experiment-capture-ambiguity}
In our main experiment, we aim to judge whether is is possible to learn how to capture the human agreement distribution. We first obtain a base model in the same manner as \citet{nie-etal-2020-learn}, by pre-training it for 3 epochs on the gold-labels of the SNLI and MNLI training sets. We observed that this pre-training step is necessary to provide the model with a general understanding of the NLI task to compensate for the low amount of ambiguity data available.
%to be an important step especially for MNLI, where performance drops significantly if we do not pre-train. Due to the low amount of ambiguity data available, this procedure is necessary to provide the model with a general understanding of the NLI task. 
We then finetune the model on our AmbiNLI dataset, setting the training objective to be the minimization of the cross-entropy between the output probability distribution and the target ambiguity distribution. For evaluation, we compute the ChaosNLI divergence scores, measured using the Jensen-Shannon Divergence (JSD), as was done in their original experiments. Furthermore, we explore what effect our ambiguity learning has on accuracy by comparing models trained on exactly the same data but with gold-label training versus ambiguous training. In order to achieve this, we prepare a version of AmbiNLI where we replace the ambiguity distributions with gold-labels. Since the two models have seen the exact same data, performance differences can be directly attributed to the process of capturing ambiguity. We report accuracy on ChaosNLI using their re-computed gold-labels.

\paragraph{Further accuracy analysis.}
To reinforce our hypothesis that accuracy improvements can be gained by leveraging the extra knowledge that models capture with ambiguity, we run an additional experiment on the ChaosMNLI dataset. We split it into three folds, and perform three-fold cross validation by training the model on two folds and evaluating on the third. Again, we start with our baseline model and compare the gold-label approach against ours.

% \paragraph{Softmax with temperature?}
% We test the effect of different temperature levels in the softmax output layer...

\paragraph{Performance in different entropy ranges.}
We also study the model performance in different entropy ranges of the ChaosMNLI set. We bin the evaluation samples based on their entropy value into three equally sized ranges, and compare the model performance on each. This experiment analyzes if the model is able to perform well in both unambiguous and highly ambiguous settings.

\paragraph{Transfer learning.} \label{transfer-learning}
In this last experiment, we aim to compare the usefulness of the representations that the BERT encoder is able to learn when training on ambiguity distributions as opposed to gold-labels. We use UNLI and IMBD movie reviews \cite{maas-etal-2011-learning} as the two downstream tasks for evaluation. As we want to focus on the representations learned during the ambiguity training phase, during the downstream task finetuning we freeze the BERT layers and update only the new classification head. We try with 1-layer and 2-layer heads using the ELU \cite{Clevert2016FastAA} activation function and a hidden size of 128. We use the original train, development and test splits for UNLI, and an 80/10/10\% split for IMDB movie reviews. We track development set loss and stop after two epochs without improvement. Each experiment is ran for 5 trials with different seeds and the mean and standard deviation are reported for each metric.

\begin{table}[]
\small
\centering
\begin{tabular}{@{}rrr@{}}
\toprule
\textbf{Folds} & \textbf{AmbiSM Gold} & \textbf{AmbiSM} \\ \midrule
0              & 0.4371               & \textbf{0.4409} \\
1              & \textbf{0.5760}      & 0.5591          \\
2              & 0.4897               & \textbf{0.5629} \\ \midrule
Average        & 0.5009               & \textbf{0.5210} \\ \bottomrule
\end{tabular}
\caption{Model accuracy when performing three-fold cross validation of a BERT base model on ChaosMNLI.}
\label{tab:mnli-folds}
\end{table}

\begin{table}[]
\small
\centering
\begin{tabular}{@{}lrr@{}}
\toprule \textbf{Entropy Range} & \textbf{JSD} & \textbf{Accuracy} \\ \midrule
Full Range & 0.2619 & 0.5810 \\
{[}0.08 - 0.58{]} & 0.2613 & \textbf{0.6706} \\
{[}0.58 - 1.08{]} & \textbf{0.2472} & 0.6262 \\
{[}1.08 - 1.58{]} & 0.2693 & 0.5087 \\
\bottomrule
\end{tabular}
\caption{\label{entropy-table} Entropy range performance comparison of the AmbiSM model.}
\end{table}

\section{Results and Discussion}

\paragraph{Training on the ambiguity distribution can reduce divergence scores.}
Table~\ref{divergence-table} details the results of our main experiment. Accuracy and JSD are provided for both the SNLI and MNLI sections in ChaosNLI. Due to differences in hyper-parameters or random seeds, we were not able to exactly reproduce the base model provided in \citet{nie-etal-2020-learn}, but achieve similar results. We follow with models further finetuned on different configurations of our AmbiNLI dataset. AmbiSM refers to the data originating from the original 5 label distribution only, while AmbiU refers to the data we obtained from UNLI. AmbiSMU thus refers to the full dataset. For each combination, we also trained a model on gold-labels (marked as ``Gold" in the table) for comparison. With the exception of ChaosSNLI when including UNLI data, every experiment has yielded a mentionable divergence score improvement. The AmbiSM model shows a 20.5\% and 21.7\% JSD decrease in ChaosSNLI and ChaosMNLI respectively. This means that we can learn to capture the human agreement distribution when we use it as a training target.

\paragraph{UNLI's skewed distribution worsens scores.}
When looking at the AmbiU and AmbiSMU results in Table~\ref{divergence-table}, it becomes apparent that UNLI data is not always beneficial. Specifically, it seems to worsen scores in all metrics except for ChaosMNLI accuracy. The distribution of labels in UNLI is drastically different from that of the remaining data, and we believe that when a model is finetuned on it, this distribution shift has a negative influence. We have found a very large number of samples with labels very close to 0 or 1, which translate into very extreme non-ambiguous distributions when converted. To confirm this, we filtered out all UNLI samples that had a probability label $p < 0.05$ or $p > 0.97$, and ran the ``Filtered" experiments. Indeed, in AmbiU, this naive filtering process yields about 20\% lower JSD scores and about 5\% higher accuracy. We conclude that UNLI data, under the current conversion approach, is somewhat problematic.

\paragraph{Training on the ambiguity distribution can yield accuracy improvements.}
We have found that, for the case of AmbiSM, a model trained to target the ambiguity distribution achieves higher accuracy. This means that more precise knowledge can be acquired when learning the true underlying ambiguity of questions instead of the sometimes misleading gold-label. When using UNLI data (AmbiU and AmbiSMU) however, the results are mixed, as discussed above. Thus, to further strengthen our argument on the benefit of ambiguity data, we refer to the supplementary experiment results in Table~\ref{tab:mnli-folds}, where we obtain a 2.1\% accuracy improvement when performing three-fold cross-validation on the ChaosMNLI dataset. 
%Finally, when performing a qualitative analysis on the predictions of AmbiSM and AmbiSM Gold, we found that the former is more accurate with neutral labels. 
When performing a qualitative analysis on the predictions of the AmbiSM and AmbiSM Gold models, we found that the former has a stronger tendency towards neutrality, both in the number of correctly predicted neutral labels and in the average neutrality score given. However, it also resulted in some examples now being incorrectly labeled as neutral. It seems to be the case that the neutral label is the main source of ambiguity. Most ambiguous questions have a considerable amount of neutral probability, which likely produces the shift.
For more details, including label counts for correct predictions as well as some prediction examples, refer to Appendix \ref{section:qualitative-analysis}.

\paragraph{Divergence scores are stable.}
Through the entropy range comparison of Table~\ref{entropy-table} we learn that divergence scores remain similar across different entropy subsets, showing that the model is capable of recognizing which questions are ambiguous, and appropriately adjusting the entropy level of its output. Accuracy dramatically decreases in high entropy ranges, but this goes along with our intuition, since both human annotators and the models will have doubts regarding the correct answer to the question, which leads to mismatches between the model prediction and the assigned label.

\paragraph{Ambiguity models learn better representations for transfer learning.}
Lastly, in Table~\ref{tab:transfer}, we observe a consistent performance improvement in transfer learning to different tasks. From the results we can infer that, by targeting the ambiguity distribution, the model can capture better linguistic representations than by targeting gold-labels. We believe that a similar trend should be visible in other tasks as well, and that the margins of improvement should increase with more ambiguity data to train on.

\paragraph{Is ambiguity training worth the extra labeling cost?}
One argument against this method is the apparent extra labeling cost required. Indeed, when comparing the gold-label and ambiguity approaches at equal number of total labels, the gold-label approach would likely attain higher performance due to the difference in number of samples. However, we argue that collecting multiple labels has several benefits other than ambiguity distribution generation. Most importantly, they help avoid mis-labelings and raise the overall quality of the dataset. In many occasions, multiple labels are already being collected for these reasons, but occasionally not released (for example, \citet{bowman-etal-2015-large} didn't release the multiple labels they collected for 10\% of the training data). They can also be used in other methods such as item response theory \cite{lalor-etal-2016-building}. Furthermore, this paper's main intention is not to encourage multi-label collection at the cost of sample quantity, but rather to show the benefits of exploiting the ambiguity distribution if it is available.

\begin{table}[]
\centering
\small
\resizebox{\linewidth}{!}{%
\begin{tabular}{@{}lrrrr@{}}
\toprule
\multicolumn{1}{c}{\textbf{}} &
  \multicolumn{2}{c}{\textbf{UNLI}} &
  \multicolumn{2}{c}{\textbf{IMDB}} \\
\multicolumn{1}{c}{\textbf{Model}} &
  \multicolumn{1}{c}{\textbf{Pearson$\uparrow$}} &
  \multicolumn{1}{c}{\textbf{MSE$\downarrow$}} &
  \multicolumn{1}{c}{\textbf{CE Loss$\downarrow$}} &
  \multicolumn{1}{c}{\textbf{Acc.$\uparrow$}} \\ \midrule
1 Layer     &                     &                     &                     &                     \\
AmbiSM G & .6331(0.9)          & .0758(0.5)          & .4727(1.7)          & .7758(6.4)          \\
AmbiSM      & \textbf{.6354(1.0)} & \textbf{.0754(0.4)} & \textbf{.4701(1.5)} & \textbf{.7783(6.1)} \\ \midrule
2 Layers    &                     &                     &                     &                     \\
AmbiSM G & .6266(5.9)          & .0765(1.0)          & .4431(0.8)          & .7906(4.3)          \\
AmbiSM      & \textbf{.6290(4.1)} & \textbf{.0762(0.7)} & \textbf{.4392(1.2)} & \textbf{.7939(3.3)} \\ \bottomrule
\end{tabular}
}
\caption{Transfer learning comparison on UNLI and IMDB movie reviews (\textit{std} is $\times 10^{-4}$). For UNLI we measure the Pearson correlation and mean squared error (MSE), following \citet{chen-etal-2020-uncertain}. For IMDB, we measure the accuracy and cross-entropy (CE) loss on the test set. \textit{G} means Gold.}
\label{tab:transfer}
\end{table}

\section{Conclusion}
We hypothesized that the intrinsic ambiguity present in natural language datasets can be exploited instead of treating it like noise. We used existing data to generate ambiguity distributions for subsets of SNLI, MNLI, and UNLI, and trained new models that are capable of more accurately capturing the ambiguity present in these datasets. Our results show that it is indeed possible to exploit this ambiguity information, and that for the same amount of data, a model trained to recognize ambiguity shows signs of higher performance in the same task as well as in other downstream tasks. 

However, our dataset was created using existing resources and lacks in quality and quantity. While it was enough to show that this research direction is promising, it limited the strength of our results. In future work, we wish to obtain larger amounts of data by using crowdsourcing techniques, and expand our scope to other NLP tasks as well.

\section*{Acknowledgments}
This work was supported by JSPS KAKENHI Grant Number 21H03502, JST PRESTO Grant Number JPMJPR20C4, and by the ``la Caixa'' Foundation (ID 100010434), under agreement LCF/BQ/AA19/11720042.

% \bibliography{references}
\bibliography{anthology,acl2021}
\bibliographystyle{acl_natbib}

\clearpage

\appendix

% appendix new section here
\section{Conversion Function} \label{section:conversion-function}
\begin{figure}
    \centering
    \includegraphics[scale=0.55]{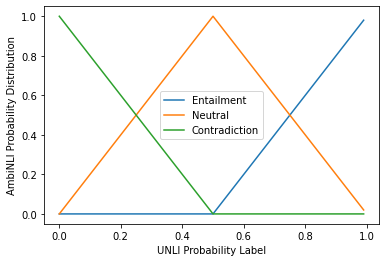}
    \caption{Linear approach to converting the UNLI regression value into an ambiguity distribution.}
    \label{fig:unli-conversion}
\end{figure}

Figure \ref{fig:unli-conversion} shows a plot of the linear conversion approach that we have taken to convert UNLI data into a probability distribution.

\section{Qualitative Analysis}\label{section:qualitative-analysis}

\renewcommand{\arraystretch}{1.5}
\begin{table*}[]
\centering
% \small
\begin{tabular}{m{0.3\linewidth}m{0.3\linewidth}m{0.08\linewidth}m{0.08\linewidth}m{0.08\linewidth}}
\toprule
\textbf{Premise} & \textbf{Hypothesis} & \textbf{CHAOS} & \textbf{ASM} & \textbf{ASMG}\\\hline
\textbf{Only AmbiSM is correct} \\\hline
They were in rotation on the ground grabbing their weapons. & The woman rolled and drew two spears before the horse had rolled and broken the rest. & $\mathrm{E}^{0.33}$ $\mathbf{N^{0.51}}$ $\mathrm{C}^{0.16}$ & $\mathrm{E}^{0.178}$ $\mathbf{N^{0.522}}$ $\mathrm{C}^{0.300}$ & $\mathrm{E}^{0.065}$ $\mathrm{N}^{0.282}$ $\mathbf{C^{0.653}}$\\\hline
Some of the unmet needs are among people who can pay, but who are deterred from seeking a lawyer because of the uncertainty about legal fees and their fear of the profession. & Some people can't afford it. & $\mathbf{E^{0.47}}$ $\mathrm{N}^{0.40}$ $\mathrm{C}^{0.13}$ & $\mathbf{E^{0.572}}$ $\mathrm{N}^{0.398}$ $\mathrm{C}^{0.030}$ & $\mathrm{E}^{0.476}$ $\mathbf{N^{0.494}}$ $\mathrm{C}^{0.030}$\\\hline
This number represents the most reliable, albeit conservative, estimate of cases closed in 1999 by LSC grantees. & This is an actual verified number of closed cases. & $\mathrm{E}^{0.21}$ $\mathrm{N}^{0.12}$ $\mathbf{C^{0.67}}$ & $\mathrm{E}^{0.281}$ $\mathrm{N}^{0.151}$ $\mathbf{C^{0.568}}$ & $\mathbf{E^{0.485}}$ $\mathrm{N}^{0.123}$ $\mathrm{C}^{0.391}$ \\\hline
\multicolumn{2}{l}{\textbf{Only AmbiSM Gold is correct}} \\\hline
And it needs work too, you know, in case I have to jump out with this parachute from my lil' blue sports plane for real.' & It needs to work Incase he has to jump out a window. & \enlidist{0.44}{0.28}{0.28} & \nnlidist{0.414}{0.429}{0.156} & \enlidist{0.489}{0.386}{0.125}\\\hline
uh wasn't that Jane Eyre no he wrote Jane Eyre too & Was it Jane Eyre or not? & \enlidist{0.58}{0.36}{0.06} & \nnlidist{0.398}{0.422}{0.180} & \enlidist{0.474}{0.413}{0.113}\\\hline
Thus, the imbalance in the volume of mail exchanged magnifies the effect of the relatively higher rates in these countries. & There is an imbalance in ingoing vs outgoing mail. & \enlidist{0.60}{0.35}{0.05} & \nnlidist{0.400}{0.499}{0.101} & \enlidist{0.458}{0.450}{0.092}\\
\bottomrule
\end{tabular}
\caption{Example of ChaosMNLI prediction for AmbiSM and AmbiSM Gold. \textbf{CHAOS} is the human distribution, \textbf{ASM} is the predicted distribution by AmbiSM and \textbf{ASMG} is the predicted distribution by AmbiSM Gold. The labels E, N, and C stand for entailment, neutral, and contradiction and their probabilities are appended.}
\label{tb:example-table}
\end{table*}

\begin{figure*}
    \centering
    \includegraphics[width=\linewidth]{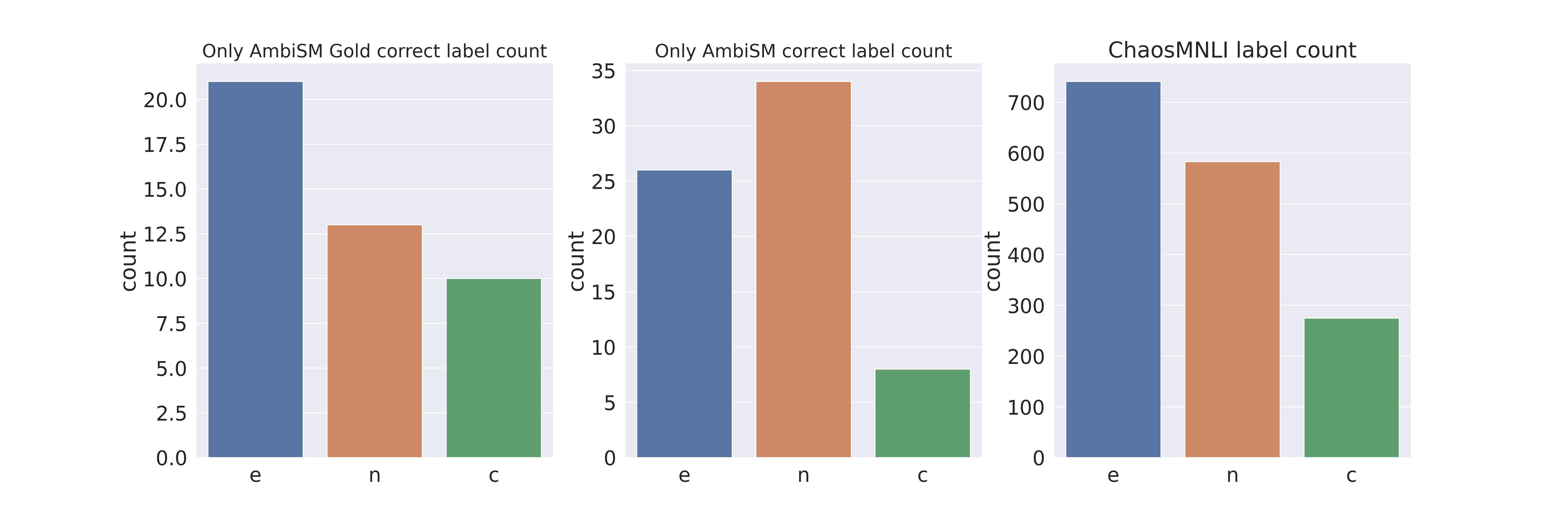}
    \caption{The count plot of the labels of the correctly predicted samples by either AmbiSM Gold or AmbiSM, AmbiSM Gold (left), AmbiSM (middle), and the labels of the whole ChaosMNLI (right). The labels e, n, and c stand for entailment, neutral, and contradiction respectively.}
    \label{fig:correct_plot}
\end{figure*}

To investigate the prediction differences between an ambiguous model and one trained on gold-labels, we compared AmbiSM and AmbiSM Gold predictions on the ChaosMNLI dataset (see Table~\ref{tb:example-table}). We use the new labels obtained from the ChaosNLI majority vote, instead of the original MNLI labels. We focus on two situations: 1) when only AmbiSM can predict the label correctly and 2) when only AmbiSM Gold can predict the label correctly. We picked samples from the high entropy regions to observe how the models deal with ambiguity. Generally, AmbiSM has a higher tendency towards neutrality. However, it was also able to show confidence in some samples that are more entailed or contradicted. On the other hand, we also observe some samples that were missed by AmbiSM due to its tendency, while AmbiSM Gold could predict them correctly.

Furthermore, we show the label counts for the samples that were correctly labeled by only one of the two models in Figure~\ref{fig:correct_plot}. The labels of the samples that are predicted correctly by AmbiSM Gold show the same distribution as the  ChaosMNLI dataset as a whole. However, within the samples that are only predicted correctly by AmbiSM we can find a higher amount of neutral labels. This emphasizes that the behavior of the model trained on ambiguity targets can deal with neutral labels in NLI better; 
neutral labels are likely to be the biggest source of ambiguity.

\end{document}